\documentclass[10pt,twocolumn,letterpaper]{article}

\usepackage{cvpr}      

\usepackage{cuted} 

\usepackage{currfile} 

\usepackage{caption} 

\usepackage{multirow}

\definecolor{cvprblue}{rgb}{0.21,0.49,0.74}
\usepackage[pagebackref,breaklinks,colorlinks,allcolors=cvprblue]{hyperref}

\def\confName{CVPR}
\def\confYear{2027}

\newcommand{\indicator}{\mathbf{1}}
\newcommand{\ours}{CFT}

\title{Adaptation Circuits for Compute-Efficient Fine-Tuning}

\author{Uri Z. Kialy\\
School of Computer Science, Ariel University, Israel\\
{\tt\small uri.kialy@msmail.ariel.ac.il}
\and
Gil Ben-Artzi\\
School of Computer Science, Ariel University, Israel\\
{\tt\small gilba@ariel.ac.il}
}

\begin{document}
\maketitle

\begin{abstract}
Parameter-Efficient Fine-Tuning (PEFT) has become the de facto standard for adapting Vision Transformers (ViTs) to downstream tasks. While parameter count has been the dominant efficiency metric in PEFT, it does not imply compute efficiency: parameter-sparse methods can still incur full-model training cost per step, and typically need long schedules to reach peak accuracy. Across seven fine-tuning methods on ViT-B/16, trainable parameters span $0.1\%$ to $100\%$ of the backbone, while the measured time of a training epoch stays within $0.74\times$ to $1.69\times$ that of full fine-tuning and does not necessarily grow with the trainable fraction. We introduce Circuit Fine-Tuning (\ours{}), which uses attribution to select an adaptation circuit, a parameter-budgeted set of attention heads and MLP blocks, before any task head is trained, and tunes it in place, reducing the number of training steps rather than the cost of a step. \ours{} adds no parameters, uses no learning-rate warmup, and leaves the inference graph unchanged. On VTAB-1k, \ours{} is comparable with the most accurate baseline while reaching its best checkpoint after $24$ epochs on average, against $35$ to $72$ for the baselines; at equal accuracy the most efficient baseline needs $1.1$ to $3.6\times$ the training FLOPs of \ours{} and $1.6$ to $3.8\times$ its training time. In a $5$-epoch comparison under domain shift, on mammography lesions (CBIS-DDSM), \ours{} has the highest F1 on ViT-B/16 and Swin-v2-B and ties on DINOv2. A feasibility study applies \ours{} to a vision-language model (Gemma-3) as well.
\end{abstract}

\section{Introduction}
\label{sec:intro}

The emergence of large-scale Vision Transformers (ViTs)~\cite{dosovitskiy2020image} has made efficient adaptation a central problem in computer vision. Full fine-tuning, which updates all parameters on a downstream task, is the natural default but becomes computationally expensive as models scale. This has driven the rapid growth of Parameter-Efficient Fine-Tuning (PEFT), where the goal is to adapt a frozen backbone by updating as few parameters as possible. Additive and modulation methods~\cite{houlsby2019parameter,jia2022visual,hu2021lora,lian2022scaling} insert new learnable modules into a frozen backbone; sensitivity-based methods~\cite{he2023sensitivity,zhang2024gradient} select which existing parameters to update. Both strategies target the same axis: parameter count. When carefully tuned, PEFT methods reach similar accuracy on VTAB-1k~\cite{mai2025lessons}. We ask a complementary question: how much compute does a method need to reach that accuracy?

\begin{figure}[t]
\centering
\includegraphics[width=\linewidth]{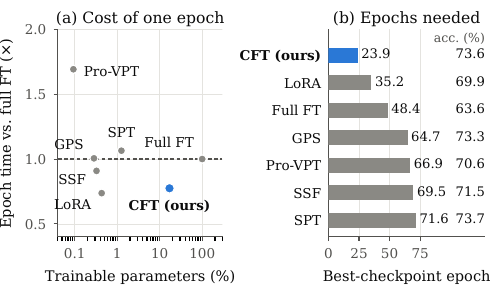}
\caption{\textbf{What an epoch costs, and how many are needed} (ViT-B/16, VTAB-1k). \textbf{(a)} Measured time per epoch relative to full fine-tuning, against the fraction of the backbone each method trains (Tab.~\ref{tab:cost}). The cost of an epoch does not necessarily grow with the trainable fraction (Spearman $\rho=-0.43$ on unrounded fractions, $n{=}7$). \textbf{(b)} Mean epoch of each method's best logged checkpoint within $100$ epochs (bar and first number), and the accuracy at that checkpoint (right column); both are the $19$-task values of Tab.~\ref{tab:vtab}.}
\label{fig:cost_vs_params}
\end{figure}

\paragraph{Parameter count is not training cost.} Parameter count is not a good proxy for training effort. We profiled seven fine-tuning methods on ViT-B/16 (Fig.~\ref{fig:cost_vs_params}a, Tab.~\ref{tab:cost}). Their trainable fraction ranges over three orders of magnitude, from $0.1\%$ to $100\%$, but the time of a training epoch ranges only from $0.74\times$ to $1.69\times$ that of full fine-tuning, and its FLOPs from $68\%$ to $151\%$. A method that updates $0.3\%$ of the weights can cost as much per epoch as full fine-tuning, and one that updates $1.2\%$ has a longer epoch. With the cost of an epoch confined to this range, how many epochs a method needs matters as much as what each epoch costs: the six baselines we evaluate reach their best checkpoint after $35$ to $72$ epochs on VTAB-1k (Fig.~\ref{fig:cost_vs_params}b).

\paragraph{Training compute as the target.}
Prior work that looks beyond parameter count reduces the cost of a step, by training a side network without back-propagating through the backbone~\cite{sung2022lst,mercea2024losa,tang2024last}, or reduces inference cost by skipping tokens~\cite{zhao2024dyt}. We take the complementary route: keep the model and its inference graph untouched, and ask which pretrained weights to update so that fewer steps are needed while keeping the cost of each epoch competitive. We refer to the resulting trade-off as \emph{compute-to-accuracy}: the accuracy a method reaches for a given amount of training compute.

\paragraph{Adaptation circuits.}
A pretrained model adapts more readily to a target distribution that is close to its pretraining distribution, but a fixed checkpoint gives no control over that. Work on mechanistic interpretability~\cite{olah2020zoom,elhage2021mathematical} shows that Transformers compute with sparse subgraphs, or circuits, which are usually identified in a trained model to explain its behavior using attribution methods. Inspired by circuit discovery approaches, we hypothesize that efficient adaptation can be obtained by updating only the subgraph of the pretrained model with the highest attribution scores on the target data. We define the set of attention heads and MLP blocks selected for updating, scored by an attribution method before any task head is trained, as an \emph{adaptation circuit}. We make no claim that it is a unique or causal explanation of the model.

\paragraph{Circuit Fine-Tuning (\ours{}).}
\ours{} has two phases. \emph{Discovery} scores every module with an attribution method from the circuit literature, applied against an untrained probe on a small calibration set, and selects a parameter-budgeted set of modules. \emph{Adaptation} fine-tunes the pretrained weights of the selected modules and freezes the rest. Sensitivity methods ask which parameters most influence the loss through their own gradients; \ours{} asks which modules' activations matter for the target data.

\paragraph{Contributions.}
\begin{enumerate}
\item \textbf{Compute-efficient fine-tuning as target.} We demonstrate that parameter count, the dominant PEFT efficiency metric, is decoupled from training compute. For seven methods we measure FLOPs and time per training epoch (Sec.~\ref{sec:cost}) and find that training compute does not necessarily grow with the trainable fraction of parameters. We advocate \emph{compute-to-accuracy} as an efficiency metric for transformer adaptation.
\item \textbf{\ours{}.} We introduce Circuit Fine-Tuning (\ours{}), a method that reaches a given accuracy with less training compute than the PEFT baselines we evaluate. \ours{} employs module-level attribution-based selection before any task head is trained, followed by in-place tuning of the selected pretrained weights. It adds no parameters, uses no warmup, and leaves the inference graph unchanged (Sec.~\ref{sec:method}).
\item \textbf{Extensive evaluation.} We evaluate \ours{} at equal epochs, equal accuracy and equal epoch budgets, and present an analysis of the selection. On VTAB-1k, \ours{} is the most accurate method at every common checkpoint up to $35$ epochs, reaches its best checkpoint after $24$ epochs on average, against $35$ to $72$ for the baselines, and reaches a given accuracy with fewer training FLOPs than any baseline (Sec.~\ref{sec:main_results}); with baselines retuned for short schedules or set to a matched parameter budget, none exceeds it, and at the matched parameter budget it leads sensitivity-based selection (Sec.~\ref{sec:budgets}). We compare the selection with sensitivity-based selection, and report what the attribution score does and does not measure (Sec.~\ref{sec:selection}). Beyond VTAB-1k, we report a comparison on CBIS-DDSM with three backbones and a feasibility study on a vision-language model.
\end{enumerate}

\section{Related Work}
\label{sec:related}

\textbf{Parameter-efficient fine-tuning.}
Additive methods insert new modules: Adapters~\cite{houlsby2019parameter}, VPT~\cite{jia2022visual}, AdaptFormer~\cite{chen2022adaptformer}, and prompt-relocation methods such as Pro-VPT~\cite{shang2025pro}. Modulation methods reparameterize or rescale: LoRA~\cite{hu2021lora}, BitFit~\cite{ben-zaken-etal-2022-bitfit}, SSF~\cite{lian2022scaling}, FacT~\cite{jie2023fact}. Recent work continues to push parameter and memory budgets~\cite{shen2024snell,tokencoord2025,fivepercent2025,colora2025}.

\textbf{Selecting pretrained weights before training.}
A separate line fixes, before or early in training, a subset of existing weights to update: FISH Mask~\cite{sung2021fish} by Fisher information, Diff Pruning~\cite{guo2021parameter} by a learned sparse difference, Child-Tuning~\cite{xu2021childtuning} by a task-driven child network, and AdaLoRA~\cite{zhang2023adalora} by adaptive parameter-budget allocation. In vision, SPT~\cite{he2023sensitivity} and GPS~\cite{zhang2024gradient} rank parameters by gradient sensitivity, TR-PTS~\cite{luo2025trpts} combines Fisher-based parameter selection with token reduction, and Surgical Fine-Tuning~\cite{lee2023surgical} chooses blocks by the type of distribution shift. The setting is close to pruning at initialization~\cite{lee2019snip}. \ours{} differs from these methods in its signal, an activation-level clean-versus-corrupted attribution computed before a task head exists, in its granularity, whole heads and MLP blocks, and in the quantity it is evaluated on, compute-to-accuracy.

\textbf{Training cost beyond parameter count.}
Ladder Side-Tuning~\cite{sung2022lst}, LoSA~\cite{mercea2024losa} and LAST~\cite{tang2024last} observe that parameter-efficient methods still back-propagate through the backbone, and avoid it with a side network, which lowers time and memory per step. Dynamic Tuning~\cite{zhao2024dyt} argues that adaptation should also be judged by computation and reduces inference FLOPs by dispatching tokens. We share the premise and act on a different term: the number of steps, with in-place updates and no architectural change. The approaches are complementary.

\textbf{Circuits and attribution.}
Circuits~\cite{olah2020zoom,elhage2021mathematical} were first found by activation patching~\cite{conmy2023automated}. Edge Attribution Patching (EAP)~\cite{syed2023attribution} approximates patching with gradients, and EAP-IG~\cite{hanna2024have} adds integrated gradients~\cite{sundararajan2017axiomatic}. These tools explain trained models; we use the attribution score, at the level of nodes, as a selector applied before fine-tuning.

\section{Method}
\label{sec:method}

\ours{} first selects an adaptation circuit (Sec.~\ref{sec:discovery}, \ref{sec:select}) and then fine-tunes it (Sec.~\ref{sec:adaptation}).

\subsection{Scoring modules before training}
\label{sec:discovery}

\paragraph{Nodes.}
Each Transformer block $\ell\in\{1,\dots,L\}$ contributes $H$ attention-head nodes and one MLP node. ViT-B/16 ($L{=}12$, $H{=}12$) has $156$ candidate nodes. For the hierarchical Swin-v2-B, stage $s$ contributes $\text{blocks}_s\cdot(\text{heads}_s{+}1)$ nodes, $400$ in total.

\paragraph{Attribution.}
For a clean target image $x$ and a corrupted baseline $x'$, obtained by randomly permuting the $16{\times}16$-pixel patches of $x$,
the EAP-IG score~\cite{hanna2024have} of node $m$ approximates the effect of patching its activation $a_m$:
\begin{equation}
\label{eq:eapig}
\phi_m(x) = \bigl(a_m(x') - a_m(x)\bigr) \cdot \frac{1}{S} \sum_{s=1}^{S}
\frac{\partial F\!\left(x' + \tfrac{s}{S}(x-x')\right)}{\partial a_m},
\end{equation}
where $F$ is a scalar output and $S$ the number of integration steps. EAP-IG is defined on edges; we use it at the level of nodes and do not recover edges or connectivity. Of three attribution methods compared on CBIS-DDSM at a $15\%$ parameter budget, EAP-IG gave the highest downstream accuracy by a small margin ($66.1$, against $65.9$ for EAP~\cite{syed2023attribution} and $65.7$ for relevance patching \cite{jafari2025relpfaithfulefficientcircuit}), and we use it throughout.

\begin{table}[t]
\centering
\scriptsize
\setlength{\tabcolsep}{4pt}
\caption{\textbf{Probe initialization} (subset of CIFAR-100, ViT-B/16, same parameter budget in all rows; mean $\pm$ std over seeds).} 
\label{tab:factorial}
\begin{tabular}{llcc}
\toprule
\textbf{Selection} & \textbf{Probe init.} & \textbf{Top-1 (\%)} & $\boldsymbol{\Delta}$ \\
\midrule
Random  & Kaiming   & 79.2$_{\pm0.8}$ & --- \\
Random  & Near-zero & 79.8$_{\pm0.7}$ & $+0.6$ \\
EAP-IG  & Kaiming   & 86.1$_{\pm0.5}$ & $+6.9$ \\
\textbf{EAP-IG (\ours{})} & \textbf{Near-zero} & \textbf{90.4}$_{\pm0.3}$ & $\mathbf{+11.2}$ \\
\bottomrule
\end{tabular}
\end{table}

\paragraph{An untrained probe.}
Scoring happens before any task head is trained, so the attribution needs a scalar output to differentiate. We replace the pretrained head with a linear probe $W_{\text{probe}}\in\mathbb{R}^{d\times C}$ drawn from $\mathcal{N}(0,\sigma^2)$ with $\sigma{=}10^{-5}$, and take $F$ to be the log-probability difference between the true class and the other classes. 
The probe is a fixed random readout of the features: what differs across modules is how strongly a clean-versus-corrupted change in their activations moves this readout on target images. We treat the score as a practical selection rule and examine it empirically (Tab.~\ref{tab:factorial}). The probe is discarded after discovery.

\paragraph{Calibration set.}
We sample $N{=}32$ training images with stratified class coverage and aggregate
\begin{equation}
\label{eq:global_score}
\Phi_m = \frac{1}{N} \sum_{i=1}^{N} |\phi_m(x_i)|.
\end{equation}
We validate this empirically: on ViT/EuroSAT, accuracy saturates at
$N{=}32$ ($92.3\%$), with negligible gain ($+0.1$) from doubling to $N{=}64$.

\subsection{Selecting the adaptation circuit}
\label{sec:select}

Scores are normalized by node size, $\tilde{\Phi}_m = \Phi_m/|\theta_m|$, so that large modules are not favored by size alone. On CBIS-DDSM this normalization improves AUC at five parameter budgets (Tab.~\ref{tab:normalize}). The parameter budget is a fraction $k$ of the $P_{\text{total}}$ backbone parameters: $B=\lfloor k\cdot P_{\text{total}}\rfloor$. A greedy procedure builds the selected set $\mathcal{T}$: (1) take the highest-ranked head of each layer, when feasible, so that every layer is covered; (2) add the next highest-ranked nodes until the parameter budget is exhausted. The mask is $M_m=\indicator[m\in\mathcal{T}]$.

\begin{table}[t]
\centering
\scriptsize
\setlength{\tabcolsep}{5pt}
\caption{\textbf{Size-normalized scoring} (CBIS-DDSM, DINOv2): raw versus normalized node scores at several parameter budgets $k$.} 
\label{tab:normalize}
\begin{tabular}{lccccc}
\toprule
$k$ (\%) & 5 & 10 & 20 & 50 & 70 \\
\midrule
AUC (raw)  & .674 & .690 & .708 & .716 & .700 \\
AUC (norm.) & \textbf{.682} & \textbf{.710} & \textbf{.713} & \textbf{.721} & \textbf{.726} \\
\bottomrule
\end{tabular}
\end{table}

\subsection{Adaptation}
\label{sec:adaptation}

The backbone is split into $\theta_{\text{active}}=\{\theta_m: M_m{=}1\}$ and $\theta_{\text{frozen}}$. A new classification head with Kaiming initialization~\cite{he2015delving} replaces the probe, and only $\theta_{\text{active}}$ and the head are trained, with AdamW and a cosine schedule.

\paragraph{No warmup.}
\ours{} updates pretrained weights and introduces  no additional adaptation modules. On three VTAB-1k tasks, adding a 5-epoch warmup leaves its accuracy unchanged, whereas removing the 10-epoch warmup of SSF's official recipe costs SSF $7.7$ points (Tab.~\ref{tab:warmup}).

\begin{table}[t]
\centering
\scriptsize
\setlength{\tabcolsep}{6pt}
\caption{\textbf{Warmup ablation} (three VTAB-1k tasks, mean over $3$ seeds).} 
\label{tab:warmup}
\begin{tabular}{llc}
\toprule
\textbf{Method} & \textbf{Warmup} & \textbf{Acc.\ (\%)} \\
\midrule
\multirow{2}{*}{\ours{}} & None   & 76.4$_{\pm0.2}$ \\
                         & 5 ep.  & 76.4$_{\pm0.2}$ \\
\midrule
\multirow{2}{*}{SSF}     & Standard & 74.73 \\
                         & None     & 67.06 \\
\bottomrule
\end{tabular}
\end{table}

\paragraph{Inference and deployment.}
\ours{} changes only the values of existing weight tensors. Architecture, tensor shapes and sequence length are those of the original model: there are no extra tokens, layers or merge steps, optimized kernels such as FlashAttention~\cite{dao2022flashattention} apply unchanged, and the result is a standard checkpoint.

\section{What Does Training Cost?}
\label{sec:cost}

\begin{table}[t]
\centering
\scriptsize
\setlength{\tabcolsep}{5pt}
\caption{\textbf{Measured cost of one training epoch} (ViT-B/16, $1{,}000$ VTAB-1k images). TFLOPs count forward and backward operations (fvcore forward count times the counted backward-to-forward ratio). Epoch time covers data loading and training steps, without evaluation; the last column is relative to full fine-tuning. \ours{} is profiled at a parameter budget of $17\%$ ($8\%$ to $19\%$ across tasks). $^\dagger$Includes Pro-VPT's extra gradient pass per epoch. }
\label{tab:cost}
\begin{tabular}{lrrrr}
\toprule
\textbf{Method}  & \textbf{Train.\ \%} & \textbf{TFLOPs} & \textbf{Epoch (s)} & \textbf{vs.\ FT} \\
\midrule
Full FT   & 100  & 52.6 & 3.73 & $1.00\times$ \\
GPS       & 0.3  & 52.6 & 3.75 & $1.01\times$ \\
SSF       & 0.3  & 35.8 & 3.39 & $0.91\times$ \\
SPT       & 1.2  & 40.1 & 3.97 & $1.06\times$ \\
LoRA      & 0.4  & 36.0 & 2.75 & $0.74\times$ \\
Pro-VPT   & 0.1  & 79.3$^\dagger$ & 6.31 & $1.69\times$ \\
\textbf{\ours{}}  & 17 & 42.6 & 2.89 & $0.77\times$ \\
\bottomrule
\end{tabular}
\end{table}

We measure the cost of training with each method. Fig.~\ref{fig:cost_vs_params}a plots epoch time against the trainable fraction, and Tab.~\ref{tab:cost} reports forward-plus-backward FLOPs per epoch and GPU timings, each method with its own implementation and recipe.

\paragraph{The cost of an epoch does not follow the parameters updated.}
The trainable fraction varies by a factor of $1{,}000$ across methods, yet FLOPs per epoch stay between $68\%$ (SSF, LoRA) and $151\%$ (Pro-VPT) of full fine-tuning, and epoch time between $0.74\times$ and $1.69\times$; neither increases with the trainable fraction. All seven methods back-propagate through the backbone. GPS updates $0.3\%$ of the weights but computes gradients for all of them before masking, and costs as much as full fine-tuning. FLOP counts do not always show the time: SPT has $76\%$ of the FLOPs of full fine-tuning and a longer epoch, and SSF has $68\%$ of the FLOPs and an epoch only $9\%$ shorter. \ours{} costs $81\%$ of full fine-tuning's FLOPs per epoch and $0.77\times$ its time. 

\paragraph{Optimizing training compute.}
\ours{} is not the most efficient method per epoch alone: three baselines have fewer FLOPs, and LoRA has a shorter epoch.
However, the baselines require $35$ to $72$ epochs to reach their best checkpoints (Sec.~\ref{sec:main_results}), whereas \ours{} converges in just $24$ epochs on average, resulting in substantially lower overall training compute.

\paragraph{Discovery cost.}
EAP-IG with $S{=}12$ steps on $N{=}32$ images takes $S\!\cdot\!N=384$ forward--backward pairs, that is $0.38$ of a VTAB-1k epoch of $1{,}000$ pairs, and a smaller fraction on larger datasets. Measured, discovery takes $5.2$\,s, the time of $1.8$ training epochs of \ours{}. All cost totals for \ours{} include it: FLOPs at the per-image cost of a full fine-tuning step, and time as measured. 

\paragraph{Peak memory.}
\ours{} peaks at $24.8$\,GB, above full fine-tuning ($21.7$) and SSF ($23.5$) and below SPT ($29.0$). The peak is set by discovery, which caches activations along the integration path once; adaptation itself stores gradients and optimizer state for the selected modules only. \ours{} does not reduce memory, and methods built for that purpose~\cite{sung2022lst,tang2024last} use far less.

\section{Experiments}
\label{sec:experiments}

\subsection{Setup}
\label{sec:setup}

\textbf{Data.} VTAB-1k~\cite{zhai2019large}: $19$ classification tasks (Natural, Specialized, Structured) with $1{,}000$ training images each; we report the mean over the $19$ tasks. CBIS-DDSM~\cite{lee2016cbis}: benign-versus-malignant mammography, far from ImageNet. We also report five FGVC datasets (Sec.~\ref{sec:main_results}).

\textbf{Models.} ViT-B/16 pretrained on ImageNet-21k ($85.8$M parameters) is the main backbone; we also use Swin-v2-B~\cite{liu2022swin}, DINOv2 ViT-B/14~\cite{oquab2023dinov2} and Gemma-3-4B-IT~\cite{Kamath2025Gemma3T}.

\textbf{Baselines.} Full fine-tuning, SSF~\cite{lian2022scaling}, SPT~\cite{he2023sensitivity}, GPS~\cite{zhang2024gradient}, Pro-VPT~\cite{shang2025pro} and LoRA~\cite{hu2021lora}. On the $19$ VTAB-1k tasks each baseline runs its official code with the per-task configuration tuned and released by its authors, and all baselines are reported for $100$ epochs, the length of their official schedules.

\textbf{\ours{}.} $N{=}32$, $S{=}12$, $\sigma{=}10^{-5}$, AdamW, a cosine schedule over $35$ epochs, no warmup. The learning rate, regularization (dropout, weight decay, label smoothing) and parameter budget $k\in[8\%,19\%]$ are selected per task on a held-out validation split. 

\textbf{Configuration.} Every cost we report is that of one training run at the reported configuration, plus circuit discovery for \ours{}. In the six-task study of Sec.~\ref{sec:budgets}, each retuned baseline is given five learning rates per task and epoch budget.  Per-epoch costs were profiled on one NVIDIA A40 GPU; the accuracy runs used A40 GPUs, with PyTorch 2.4.1. 

\subsection{VTAB-1k}
\label{sec:main_results}

\begin{table*}[t]
\centering
\small
\setlength{\tabcolsep}{6pt}
\caption{\textbf{VTAB-1k at each method's best checkpoint} (ViT-B/16, 19 tasks, one run per task). Accuracy by group and overall at the best logged checkpoint of each run and the cost of reaching that checkpoint: mean best epoch; training TFLOPs and training time, both computed as best epoch times the per-epoch cost of Tab.~\ref{tab:cost} (\ours{} includes discovery; the training time of GPS includes its mask computation). Each task is read at its own best epoch, within $100$ epochs for the baselines and $35$ for \ours{}. Costs are to each method's own best checkpoint, at different accuracies; Tab.~\ref{tab:fixed} gives accuracy at fixed epochs and cost at equal accuracy.}

\label{tab:vtab}
\begin{tabular}{lcccc|ccc}
\toprule
\textbf{Method} & \textbf{Natural} (7) & \textbf{Specialized} (4) & \textbf{Structured} (8) & \textbf{Mean} (19) & \textbf{Best epoch} & \textbf{TFLOPs} & \textbf{Train (min)} \\
\midrule
Full fine-tuning & 72.9 & 83.9 & 45.4 & 63.6 & 48.4 & 2{,}547 & 3.0 \\
LoRA~\cite{hu2021lora} & 78.6 & 83.8 & 55.2 & 69.9 & 35.2 & 1{,}268 & 1.6 \\
SSF~\cite{lian2022scaling} & \textbf{82.4} & 86.0 & 54.8 & 71.5 & 69.5 & 2{,}487 & 3.9 \\
Pro-VPT~\cite{shang2025pro} & 80.3 & 85.1 & 54.8 & 70.6 & 66.9 & 5{,}309 & 7.0 \\
SPT~\cite{he2023sensitivity} & 81.5 & \textbf{86.2} & 60.5 & \textbf{73.7} & 71.6 & 2{,}870 & 4.7 \\
GPS~\cite{zhang2024gradient} & 81.3 & 86.1 & 59.9 & 73.3 & 64.7 & 3{,}405 & 4.2 \\
\midrule
\textbf{\ours{}} & 81.9 & 84.7 & \textbf{60.9} & 73.6 & \textbf{23.9} & \textbf{1{,}040} & \textbf{1.2} \\
\bottomrule
\end{tabular}
\end{table*}

\begin{figure*}[t]
\centering
\includegraphics[width=\linewidth]{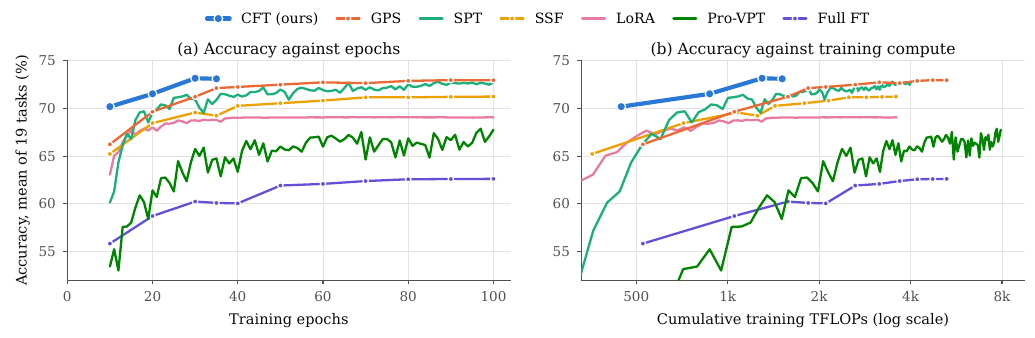}
\caption{\textbf{Accuracy against epochs and against training compute} (VTAB-1k, ViT-B/16, mean of $19$ tasks, one run per task). Baselines are shown for $100$ epochs; \ours{} trains for $35$. Markers are logged checkpoints; LoRA, SPT and Pro-VPT are logged every epoch and shown unsmoothed. Compute is the number of epochs times the measured cost of an epoch (Tab.~\ref{tab:cost}), plus circuit discovery for \ours{}.}
\label{fig:curves}
\end{figure*}

\begin{table*}[t]
\centering
\footnotesize
\setlength{\tabcolsep}{5pt}
\caption{\textbf{Accuracy at fixed epochs and cost at equal accuracy} (VTAB-1k, ViT-B/16, mean of $19$ tasks, one run per task; no checkpoint selection). Left: accuracy with all $19$ tasks read at the same epoch, hence below the per-task best checkpoints of Tab.~\ref{tab:vtab}; baselines are reported up to $100$ epochs, \ours{} trains for $35$. Right: training TFLOPs, with the epoch in parentheses, at the first logged checkpoint whose accuracy, rounded as printed, reaches the threshold (``--'': not reached).}
\label{tab:fixed}
\begin{tabular}{l rrrrrr @{\hspace{10pt}} rrrr}
\toprule
 & \multicolumn{6}{c}{\textbf{Accuracy (\%) after epoch}} & \multicolumn{4}{c}{\textbf{TFLOPs (epochs) to reach}} \\
\cmidrule(lr){2-7}\cmidrule(l){8-11}
\textbf{Method} & 10 & 20 & 30 & 35 & 50 & 100 & 70\% & 71\% & 72\% & 73\% \\
\midrule
Full FT & 55.8 & 58.7 & 60.2 & 60.1 & 61.9 & 62.6 & -- & -- & -- & -- \\
LoRA~\cite{hu2021lora} & 63.1 & 68.0 & 68.7 & 68.8 & 69.1 & 69.1 & -- & -- & -- & -- \\
SSF~\cite{lian2022scaling} & 65.2 & 68.5 & 69.6 & 69.2 & 70.5 & 71.2 & 1{,}432 (40) & 2{,}506 (70) & -- & -- \\
Pro-VPT~\cite{shang2025pro} & 53.4 & 61.4 & 65.8 & 64.8 & 66.0 & 67.8 & -- & -- & -- & -- \\
SPT~\cite{he2023sensitivity} & 60.1 & 69.6 & 70.9 & 70.9 & 72.0 & 72.7 & 882 (22) & 1{,}002 (25) & 1{,}845 (46) & -- \\
GPS~\cite{zhang2024gradient} & 66.2 & 69.7 & 71.2 & 72.1 & \textbf{72.5} & \textbf{73.0} & 1{,}578 (30) & 1{,}578 (30) & 1{,}841 (35) & 4{,}734 (90) \\
\midrule
\textbf{\ours{}} & \textbf{70.2} & \textbf{71.5} & \textbf{73.2} & \textbf{73.1} & -- & -- & 446 (10) & 872 (20) & 1{,}298 (30) & 1{,}298 (30) \\
\bottomrule
\end{tabular}

\end{table*}

\paragraph{Accuracy at the best checkpoint.}
Tab.~\ref{tab:vtab} follows the usual VTAB-1k protocol: each task is read at its own best logged checkpoint. Tab.~\ref{tab:fixed} reads all tasks at the same epoch, which gives lower numbers because tasks peak at different epochs: SPT has $73.7\%$ in Tab.~\ref{tab:vtab} and $72.7\%$ at epoch $100$; \ours{} has $73.6\%$ and $73.1\%$ (epoch $35$).

At the best checkpoint, \ours{} reaches $73.6\%$, within $0.1$ points of SPT, the most accurate baseline ($73.7\%$), and above the other five ($63.6$ to $73.3$). It has the highest accuracy on Structured tasks ($60.9$ against $60.5$ for SPT), is second on Natural tasks ($81.9$ against $82.4$ for SSF), and is below four baselines on Specialized tasks ($84.7$ against $85.1$ to $86.2$). The top three methods are within $0.4$ points of each other on a single run, a difference comparable to the variation we observe across discovery settings (Sec.~\ref{sec:selection}); we read \ours{} as matching the strongest PEFT methods in accuracy, not as exceeding them. The main advantage of \ours{} is therefore not higher peak accuracy, but reaching comparably high accuracy with less training compute (Tab.~\ref{tab:fixed}).

\paragraph{Cost to the best checkpoint.}
The best checkpoint of \ours{} arrives after $23.9$ epochs on average, against $35.2$ to $71.6$ for the baselines (Fig.~\ref{fig:cost_vs_params}b). To reach their own best checkpoints, the baselines need $1.5$ to $3.0\times$ the epochs of \ours{}, $1.2$ to $5.1\times$ its training FLOPs and $1.3$ to $5.7\times$ its training time (Tab.~\ref{tab:vtab}). These are costs to each method's own best checkpoint, at accuracies between $63.6\%$ and $73.7\%$. Parameter count does not predict them: GPS updates fewer weights than any other selection method here and costs more to its best checkpoint than full fine-tuning.

\paragraph{Equal epochs and equal accuracy.}
Tab.~\ref{tab:fixed} and Fig.~\ref{fig:curves} compare methods at the same epoch and at the same accuracy. After $10$, $20$, $30$ and $35$ epochs \ours{} is the most accurate method, by $4.0$, $1.8$, $2.0$ and $1.0$ points over the best baseline at that epoch, and at each of these checkpoints it is above every baseline that has spent the same or fewer training FLOPs (Fig.~\ref{fig:curves}b). It reaches $73.2\%$ after $30$ epochs. Among the baselines only GPS reaches $73.0\%$ within $100$ epochs, after $90$. For a common accuracy of $70$, $71$, $72$ and $73\%$, the most efficient baseline needs $2.0$, $1.1$, $1.4$ and $3.6\times$ the training FLOPs of \ours{}; full fine-tuning, LoRA and Pro-VPT do not reach $70\%$ on average. In training time, with the epoch times of Tab.~\ref{tab:cost} and the one-off costs of discovery ($5.2$\,s) and of the GPS mask ($11.9$\,s), the most efficient baseline needs $2.6$, $1.6$, $1.6$ and $3.8\times$ the time of \ours{}.

The baselines here use schedules written for long runs, which may explain part of the early gap; Sec.~\ref{sec:budgets} retunes them for short epoch budgets. 

\paragraph{Fine-grained recognition.}
On five FGVC datasets (CUB-200, NABirds, Oxford Flowers, Stanford Dogs, Stanford Cars) \ours{} reaches $88.6\%$ average accuracy with $25$ epochs of training; the baselines, trained for $100$ epochs with their published recipes, reach $84.8$ (LoRA) and $88.5$ to $89.3$. This comparison does not retune the baselines for shorter schedules.

\subsection{Equal epoch budgets with retuned baselines}
\label{sec:budgets}

\begin{table}[t]
\centering
\scriptsize
\setlength{\tabcolsep}{4pt}
\renewcommand{\arraystretch}{0.95}
\caption{\textbf{Equal budgets} (mean of six VTAB-1k tasks, ViT-B/16). Every column is a separate set of runs: the schedule is annealed over that epoch budget and the baseline learning rate is retuned. Top: each baseline at its published parameter budget (publ.; see Tab.~\ref{tab:cost}); Bottom: an adjustable parameter budget, matched to \ours{}.}

\label{tab:budgets}
\begin{tabular}{lrrrrr}
\toprule
\textbf{Method} & \textbf{Train.\ \%} & \textbf{Ep.\ 10} & \textbf{Ep.\ 20} & \textbf{Ep.\ 30} & \textbf{Ep.\ 35} \\
\midrule
\textbf{\ours{}} & ${\sim}15$ & \textbf{74.70} & \textbf{76.80} & \textbf{77.53} & \textbf{77.83} \\
\midrule
\multicolumn{6}{l}{\textit{Tuned with published parameter budget}} \\
Full FT  & publ. & 74.04 & 75.32 & 75.45 & 75.47 \\
SPT      & publ. & 73.11 & 75.53 & 76.58 & 77.05 \\
SSF      & publ. & 72.46 & 76.00 & 76.42 & 76.60 \\
LoRA     & publ. & 71.07 & 74.29 & 74.03 & 73.94 \\
GPS      & publ. & 69.71 & 73.28 & 74.90 & 75.01 \\
Pro-VPT  & publ. & 66.82 & 70.89 & 73.28 & 73.63 \\
\midrule
\multicolumn{6}{l}{\textit{Tuned with matched parameter budget}} \\
SPT      & 15.50 & 70.65 & 74.61 & 75.95 & 76.30 \\
LoRA    & 16.18  & 71.84 & 74.05 & 74.41 & 74.42 \\
GPS      & 16.15 & 72.77 & 74.94 & 75.23 & 75.23 \\
\bottomrule
\end{tabular}
\end{table}

\paragraph{Are the baselines slow only because of their recipes?}
Published recipes are written for long runs, and Tab.~\ref{tab:fixed} reads them at early epochs. Here we give the baselines a recipe built for the epoch budget they are given. We train a separate run for each epoch budget: the cosine schedule is annealed over that epoch budget, the warmup of the published recipe is kept and scaled, and the learning rate is retuned over five values around its published default, $\{\eta_{\text{def}}\!\cdot\!10^{\pm1},\eta_{\text{def}}\!\cdot\!10^{\pm1/2},\eta_{\text{def}}\}$, on a validation split. We run six tasks, two per group: DTD and Oxford-IIIT Pet, EuroSAT and RESISC45, KITTI and SmallNORB-Elevation. No baseline exceeds \ours{} at any epoch budget (Tab.~\ref{tab:budgets}, top); its margin over the best of them is $0.7$ to $1.0$ points ($74.70$ against $74.04$ for full fine-tuning at $10$ epochs, $77.83$ against $77.05$ for SPT at $35$).

\paragraph{Is it only the larger parameter budget?}
\ours{} updates about $15\%$ of the backbone, far more than the baselines. Three of the baselines have a setting for their parameter budget: the rank of LoRA and the selection budgets of SPT and GPS. We set it so that each trains a comparable fraction of the backbone and train them for the same epoch budgets (Tab.~\ref{tab:budgets}, bottom); the learning rate is selected per task. A larger parameter budget helps GPS at short schedules, by $3.1$ points at $10$ epochs, and changes LoRA by less than a point. Trainable capacity can therefore account for part of the fast convergence of \ours{}. It does not account for all of it: none of the three exceeds \ours{} at any epoch budget. Against the two sensitivity-based selectors, GPS and SPT, which also choose which pretrained weights to train, \ours{} leads by $1.5$ to $4.1$ points, and against LoRA by $2.8$ to $3.4$. These two comparisons address the two most direct alternative explanations of the faster convergence of \ours{}: recipes written for long schedules, and a larger trainable capacity. No baseline exceeds \ours{} under either, which suggests that which modules are updated contributes to its compute-to-accuracy; Sec.~\ref{sec:selection} examines the selection directly.

\subsection{What does the selection contribute?}
\label{sec:selection}

\begin{table}[t]
\centering
\scriptsize
\setlength{\tabcolsep}{3pt}
\caption{\textbf{Attribution controls} (six VTAB-1k tasks, ViT-B/16; ranges over tasks). Jaccard is the overlap of selected masks; Spearman compares the rankings of all $156$ node scores.}
\label{tab:attr_controls}
\begin{tabular}{lcc}
\toprule
\textbf{Comparison} & \textbf{Jaccard} & \textbf{Spearman} \\
\midrule
Activation shift only (no gradient) vs.\ \ours{} & 0.03--0.30 & $-$0.12--0.33 \\
Random direction vs.\ \ours{} & 0.23--0.42 & 0.32--0.74 \\
Two independent probes & 0.26--0.38 & 0.43--0.64 \\
Chance (random mask) & 0.10--0.16 & 0 \\
\bottomrule
\end{tabular}
\end{table}

Sec.~\ref{sec:budgets} compared \ours{} with existing methods at equal epoch and parameter budgets. Here we vary the selection.

\paragraph{Against the complement.}
To assess whether the selected modules provide a useful adaptation subset, we compare training them against training their complement. On CUB-200 with Swin-v2-B and a 5-epoch budget, training the selected $5\%$ of the backbone reaches $79.10\%$, while training the remaining $95\%$ reaches $77.41\%$. Despite using far fewer trainable parameters, the selected subset performs better in this setting, supporting its effectiveness for short-budget adaptation.

\paragraph{Against random selection.}
Tab.~\ref{tab:factorial} trains the same parameter budget selected at random or by EAP-IG on a subset of CIFAR-100. Attribution-based selection gives $6.9$ points more than random selection with a Kaiming-initialized probe and $10.6$ points more with the near-zero probe.

\paragraph{Against sensitivity-based selection.}
Besides being more accurate than GPS and SPT at a matched parameter budget (Tab.~\ref{tab:budgets}), \ours{} selects different weights. Averaged over the $19$ VTAB-1k tasks, the Jaccard overlap between the weights selected by \ours{} and by GPS is $0.006$ at a $1\%$ parameter budget (chance: $0.005$) and $0.153$ at $20\%$ (chance: $0.111$): at a matched $20\%$ parameter budget, about $73.5\%$ of the weights each method selects are not selected by the other. Together, the random-selection control (Tab.~\ref{tab:factorial}), the complement comparison and the matched-budget baselines (Tab.~\ref{tab:budgets}) indicate that the benefit is not explained by the number of updated parameters alone.

\paragraph{What the score measures.}
Tab.~\ref{tab:attr_controls} examines the score on six VTAB-1k tasks. First, it is not an activation-shift statistic: ranking nodes by the clean-versus-corrupted activation difference alone gives rankings uncorrelated with \ours{} (Spearman $-0.12$ to $0.33$), so the gradient term carries information. Second, the direction of the probe affects the selection, but some ranking structure persists across directions: replacing the gradient by a random direction of matched norm, or drawing a second probe, gives masks that overlap the original above chance but far from fully. The selected set is therefore not unique, and we do not interpret individual modules.

\paragraph{Stability of accuracy.}
Although masks vary with the probe, downstream accuracy is stable: on VTAB-1k the standard deviation is $\pm0.41$ over the number of integration steps and $\pm0.47$ over resampled calibration sets. Averaging the scores of up to $R{=}10$ independently drawn near-zero probes changes accuracy by at most $0.1$ points on the CIFAR-100 subset ($90.4$ to $90.5$), so we use a single probe.

\paragraph{Selected sets across tasks.}
On Swin-v2-B ($400$ candidate nodes), no node is selected for all $19$ VTAB-1k tasks; $28$ are selected for only one or two, and $6$ for twelve or more, while pairwise overlap is above the expected for independent random masks at the same parameter budget. Because masks also vary with the probe (Tab.~\ref{tab:attr_controls}), we report this as a description of the selected sets; it is the reason we run discovery per task.

\subsection{Beyond VTAB-1k}
\label{sec:medical}

\begin{table}[t]
\centering
\scriptsize
\setlength{\tabcolsep}{4.5pt}
\caption{\textbf{CBIS-DDSM at a fixed 5-epoch budget.} F1 of the malignant class (\%) on three backbones; on ViT-B/16 also accuracy (\%) and AUC.}
\label{tab:cbis}
\begin{tabular}{l ccc c c}
\toprule
 & \multicolumn{3}{c}{\textbf{ViT-B/16}} & \textbf{Swin-v2-B} & \textbf{DINOv2} \\
\cmidrule(lr){2-4}\cmidrule(lr){5-5}\cmidrule(lr){6-6}
\textbf{Method} & F1 & Acc. & AUC & F1 & F1 \\
\midrule
Full fine-tuning & 61.3 & 64.1 & 0.652 & 62.3 & 60.6 \\
SSF~\cite{lian2022scaling} & 58.0 & 62.5 & 0.626 & 64.0 & \textbf{67.4} \\
SPT~\cite{he2023sensitivity} & 61.8 & 71.9 & 0.766 & 66.1 & 67.3 \\
GPS~\cite{zhang2024gradient} & 58.4 & 60.8 & 0.695 & 64.8 & 66.4 \\
LoRA~\cite{hu2021lora} & 63.0 & 70.3 & 0.761 & 65.8 & 66.6 \\
Pro-VPT~\cite{shang2025pro} & 58.2 & 69.5 & 0.648 & 52.8 & 56.3 \\
\textbf{\ours{}} & \textbf{64.9} & \textbf{73.7} & \textbf{0.768} & \textbf{66.9} & \textbf{67.4} \\
\bottomrule
\end{tabular}
\end{table}

\paragraph{Mammography under domain shift.}
CBIS-DDSM~\cite{lee2016cbis} is small, class-imbalanced and far from ImageNet. We classify lesion crops, from mass and calcification cases together, as malignant or benign, with the official split ($2{,}863$ training and $704$ test images). Crops are converted to grayscale, resized with bicubic interpolation to $224{\times}224$ ($256{\times}256$ for Swin-v2-B) and replicated to three channels; benign-without-callback lesions are counted as benign. Every method trains for the same $5$ epochs, with the baseline learning rates retuned as in Sec.~\ref{sec:budgets}. For discovery on this dataset, $x'$ is a cutout of $x$, with a random $64{\times}64$ square set to zero, instead of the patch permutation. 
Because accuracy can be inflated by favoring the majority class, our primary metric is the F1 of the malignant class; on ViT-B/16 we also report accuracy and AUC, the usual metric in mammography. This is a comparison at a fixed epoch budget, not a converged one. \ours{} has the highest F1 on ViT-B/16 ($+1.9$ over LoRA) and Swin-v2-B ($+0.8$ over SPT) and is tied with SSF on DINOv2 (Tab.~\ref{tab:cbis}). On ViT-B/16 it also has the highest accuracy ($73.7\%$, $+1.8$ over SPT), while its AUC ($0.768$) is level with those of SPT ($0.766$) and LoRA ($0.761$). Over $3$ seeds the standard deviation is $\pm0.37$ for F1 and $\pm0.013$ for AUC, so the three AUC values cannot be separated. 

\begin{table}[t]
\centering
\scriptsize
\setlength{\tabcolsep}{8pt}
\caption{\textbf{Gemma-3-4B-IT on CUB-200}, a feasibility study. All trained methods use the same $5$ epochs.}
\label{tab:vlm}
\begin{tabular}{lcc}
\toprule
\textbf{Method} & \textbf{Updated \%} & \textbf{Top-1 (\%)} \\
\midrule
Zero-shot & 0 & 0.6 \\
Full fine-tuning & 100 & 68.0 \\
LoRA ($r{=}16$)~\cite{hu2021lora} & 0.15 & 68.3 \\
\textbf{\ours{}} & 17 & \textbf{71.3} \\
\bottomrule
\end{tabular}
\end{table}

\paragraph{A vision-language model.}
As a study beyond vision backbones, we adapt the procedure to a vision-language model, Gemma-3-4B-IT~\cite{Kamath2025Gemma3T} ($4.30$B parameters), on CUB-200 (Tab.~\ref{tab:vlm}). The model already has an output head, so no probe is attached: $F$ is the cross-entropy, at the last prompt position, of its prediction for the first token of the correct class name. In place of a corrupted image, $x'$ is a counterfactual: a training image of the class most often confused with that of $x$ in the zero-shot confusion matrix, with the same prompt. Following the circuit analysis of vision-language models by Nikankin~\etal~\cite{nikankin2025same}, the candidates are the attention heads and MLP blocks of the language model only, with nodes defined as in Sec.~\ref{sec:method}; the vision encoder and the projector stay frozen. Discovery uses $20\%$ of the training set (about $1{,}199$ images) and $8$ integration steps, and the selected set holds $729.9$M parameters, $17\%$ of the model. Normalization parameters, less than $0.005\%$ of the model, are also trained.

All methods train for $5$ epochs on the official split ($5{,}994$ training and $5{,}794$ test images) with AdamW and a cosine schedule. The learning rate is $5\times10^{-5}$ for \ours{} (effective batch size $32$), $2\times10^{-4}$ for LoRA and $2\times10^{-5}$ for full fine-tuning. LoRA uses rank $16$, $\alpha{=}32$ and dropout $0.05$ on the query and value projections of both the language model and the vision encoder ($6.45$M parameters, $0.15\%$). The image is given through the chat template with the prompt ``Classify this image into one of: [the $200$ class names]. Answer with the class name only.'' Decoding is greedy, with at most $10$ new tokens. The answer is mapped to a class by the first rule that applies: exact match; a class name contained in the answer, longest first; the answer contained in a class name; largest word overlap; a numeric fallback. An answer that matches no class counts as an error. 

Under this protocol zero-shot accuracy is $0.6\%$, so the task requires adaptation. \ours{} reaches $71.3\%$ top-1, against $68.3\%$ for LoRA and $68.0\%$ for full fine-tuning. Full fine-tuning reaches $44.8$, $64.3$, $68.0$, $67.8$ and $67.7\%$ after epochs $1$ to $5$; we report its best value. Within this epoch budget, updating all weights is no better than updating the selected $17\%$. \ours{} adds no parameters, so there is no adapter to insert or merge and the result is a standard checkpoint. We compare with LoRA and full fine-tuning only: applying SSF, SPT, GPS or Pro-VPT to a model with interleaved text tokens and a language decoder requires design choices that their papers do not specify. This is one dataset, two baselines and no compute measurement: it shows that the selection-and-update procedure carries over to a model that already has an output head, and we draw no conclusion about vision-language adaptation in general.

\begin{table}[t]
\centering
\scriptsize
\setlength{\tabcolsep}{4pt}
\caption{\textbf{Retention of the pretrained task.} ImageNet top-1 (\%) of the ViT-B/16 backbone ($80.7$ before adaptation) after adapting to three VTAB-1k tasks.}
\label{tab:retention}
\begin{tabular}{lcccc}
\toprule
\textbf{Method} & \textbf{DTD} & \textbf{EuroSAT} & \textbf{sNORB-Ele} & \textbf{Mean drop} \\
\midrule
Full fine-tuning & 42.2 & 51.3 & 0.2  & $-49.5$ \\
Pro-VPT          & 53.0 & 31.4 & 20.9 & $-45.6$ \\
LoRA             & 78.0 & 69.8 & 77.5 & $-5.6$ \\
SPT              & 76.6 & 76.8 & 76.4 & $-4.1$ \\
SSF              & 79.7 & 78.1 & 71.8 & $-4.2$ \\
\textbf{\ours{}} & 76.0 & 78.1 & 78.0 & $-3.3$ \\
GPS              & 78.5 & 79.9 & 79.6 & $\mathbf{-1.4}$ \\
\bottomrule
\end{tabular}
\end{table}

\paragraph{Retention of the pretrained task.}
In-place updates could damage the pretrained model. Tab.~\ref{tab:retention} evaluates ImageNet accuracy after adaptation. \ours{} loses $3.3$ points on average, against $49.5$ for full fine-tuning and $45.6$ for Pro-VPT, and less than LoRA, SPT and SSF; GPS, which updates far fewer weights, retains more.

\section{Conclusion}
\label{sec:conclusion}
The number of parameters a fine-tuning method updates says little about what it costs to train: across seven methods the measured cost of an epoch stays within $0.74$ to $1.69\times$ that of full fine-tuning and does not necessarily grow with the trainable fraction, which makes the number of epochs a first-order factor. Circuit Fine-Tuning selects an adaptation circuit, a parameter-budgeted set of pretrained heads and MLP blocks scored by attribution before any task head is trained, and updates it in place. On VTAB-1k it is the most accurate of seven methods at every common checkpoint up to $35$ epochs, the most efficient baseline needs $1.1$ to $3.6\times$ its training FLOPs to reach the same accuracy, and at its best checkpoint it is within $0.1$ points of the most accurate baseline. On six tasks, no baseline exceeds it at epoch budgets of $10$ to $35$, whether at its published parameter budget or at one matched to \ours{}. The model and its inference graph are left unchanged.

\section*{LLM Usage Statement}
Large language model assistants (Claude, Anthropic; ChatGPT, OpenAI) were used to help writing this paper. The authors take responsibility for the content of this paper.

{
    \small
    \bibliographystyle{ieeenat_fullname}
    \bibliography{main}
}

\end{document}